\pdfoutput=1

\documentclass[11pt]{article}

\usepackage[final]{acl}

\usepackage{times}
\usepackage{latexsym}
\usepackage{hyperref}
\usepackage{url}
\usepackage{pifont}
\usepackage{algorithmic}
\usepackage{algorithm}
\usepackage{amsmath, amssymb}
\usepackage{graphicx}
\usepackage{booktabs}
\usepackage{multirow}
\usepackage{graphicx}
\usepackage{xspace}
\usepackage{xcolor,colortbl}
\usepackage{xcolor} 
\usepackage{booktabs}
\usepackage{makecell}
\usepackage{graphicx}
\usepackage{subcaption}
\usepackage{multirow}  
\usepackage{booktabs}  
\usepackage{graphicx}  
\usepackage{xcolor}    
\usepackage{xspace}
\usepackage{colortbl}  
\usepackage[breakable,skins]{tcolorbox} 

\definecolor{red3}{RGB}{205,55,55}
\definecolor{primaryblue}{RGB}{102,126,234}
\definecolor{secondaryblue}{RGB}{79,172,254}
\definecolor{lightgray}{RGB}{245,245,245}
\definecolor{darkgray}{RGB}{64,64,64}
\definecolor{promptboxblue}{RGB}{51, 122, 183}
\usepackage{enumitem}
\usepackage{booktabs}
\tcbuselibrary{skins,breakable}

\newcommand{\ry}[1]{\textcolor{red}{ry: #1}}

\newtcolorbox{exmp}[3][]{
    enhanced,
    breakable,
    colback=lightgray,
    colframe=darkgray,
    colbacktitle=darkgray,
    coltitle=white,
    fonttitle=\bfseries,
    title={#2},
    label={#3},
    top=0.5mm,
    bottom=0.5mm,
    left=2mm,
    right=2mm,
    boxsep=0.5mm,
    toptitle=1mm,
    bottomtitle=1mm,
    attach boxed title to top left={yshift=-2mm,xshift=3mm},
    boxed title style={
        enhanced,
        size=small,
        colback=darkgray,
        colframe=darkgray,
    },
    sharp corners,
    #1
}
\newcommand{\frameworkname}{\textsc{DS\textsuperscript{2}-Instruct}}

\usepackage[T1]{fontenc}

\usepackage[utf8]{inputenc}

\usepackage{microtype}

\usepackage{inconsolata}

\usepackage{graphicx}

\title{\textsc{DS\textsuperscript{2}-Instruct}: Domain-Specific Data Synthesis \\ for Large Language Models Instruction Tuning}

\author{
  Ruiyao Xu\textsuperscript{1}\thanks{Corresponding author.} \qquad 
  Noelle I. Samia\textsuperscript{1}\thanks{Co-senior authorship.} \qquad 
  Han Liu\textsuperscript{1}\footnotemark[\value{footnote}] \\[0.5em]
  \textsuperscript{1}Northwestern University \\[0.3em]
  \texttt{ruiyaoxu2028@u.northwestern.edu}, \texttt{\{n-samia, hanliu\}@northwestern.edu}
}

\begin{document}
\maketitle
\begin{abstract}

Adapting Large Language Models (LLMs) to specialized domains requires high-quality instruction tuning datasets, which are expensive to create through human annotation. Existing data synthesis methods focus on general-purpose tasks and fail to capture domain-specific terminology and reasoning patterns. To address this, we introduce \textsc{DS\textsuperscript{2}-Instruct}, a zero-shot framework that generates domain-specific instruction datasets without human supervision. Our approach first generates task-informed keywords to ensure comprehensive domain coverage. It then creates diverse instructions by pairing these keywords with different cognitive levels from Bloom's Taxonomy. Finally, it uses self-consistency validation to ensure data quality. We apply this framework to generate datasets across seven challenging domains, such as mathematics, finance, and logical reasoning. Comprehensive evaluation demonstrates that models fine-tuned on our generated data achieve substantial improvements over existing data generation methods. \footnote{Our code is available at \url{https://github.com/rux001/DS2-Instruct}.}
\end{abstract}

\section{Introduction}
\label{sec:introduction}

Instruction-tuning Large Language Models (LLMs) requires high-quality data~\cite{wang-etal-2023-self-instruct, ouyang2022training, kopf2023openassistant, yin2023instruction, conover2023instruction}. Since acquiring large-scale datasets for fine-tuning LLMs can be prohibitively expensive due to human annotation costs, many methods use LLMs to synthesize data instead~\cite{wang-etal-2023-self-instruct,yuan2024self,xu2023wizardlm,honovich2023unnatural,xu2023instruction,mehrabi2023flacuna,ding-etal-2023-enhancing}. However, these data synthesis methods typically target general-purpose models, overlooking the need for domain-specific tuning which is critical for many real-world applications. For example, a legal chatbot must master law-related terminology, reasoning patterns, and procedures.

\begin{figure}[tbp]
\centering
\includegraphics[width=\columnwidth]{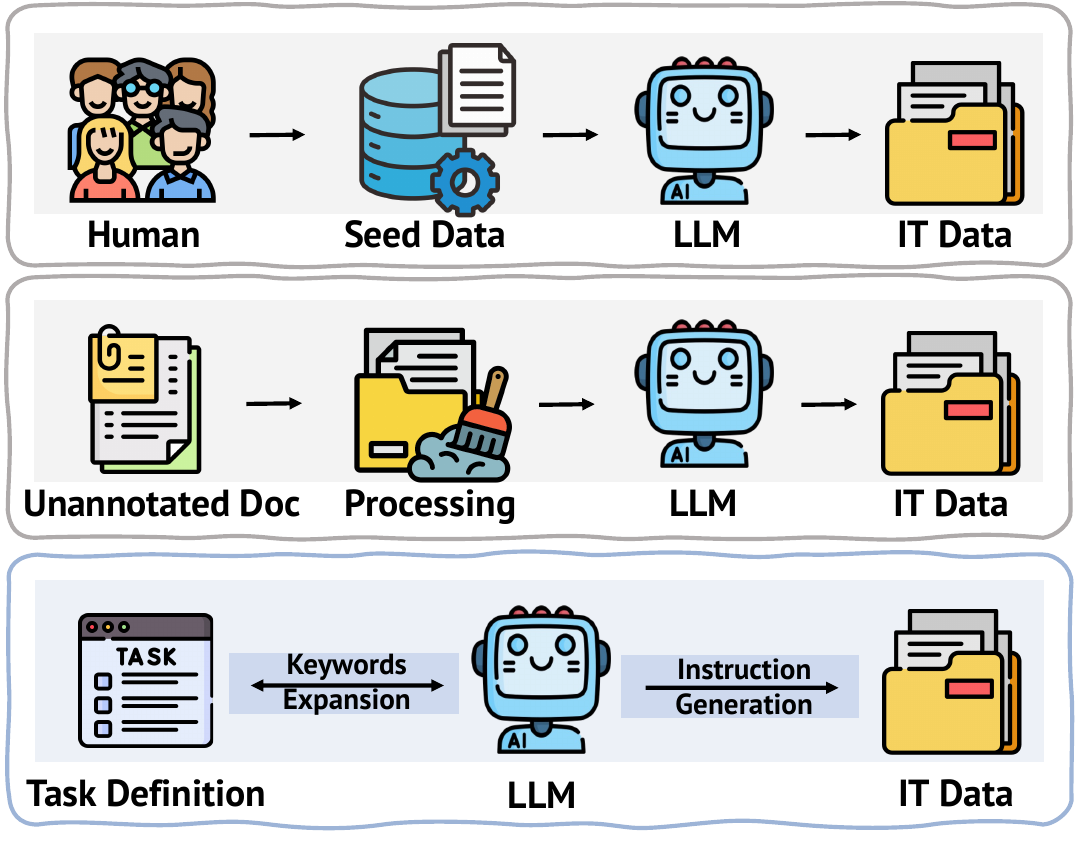}
\caption{\textbf{Top}: Traditional methods require human seed data with manual annotation. \textbf{Middle}: Document-based approaches requiring domain-specific corpora and preprocessing. \textbf{Bottom}: Our \textsc{DS\textsuperscript{2}-Instruct} framework generates data using only task definitions through systematic keywords expansion and instruction generation.}
\label{fig:framework}
\vspace{-15pt}
\end{figure}

Despite the critical importance of domain-specific instruction data, limited work has explored scalable and automatic generation methods tailored for specific tasks or domains~\cite{bonito:aclfindings24, wan-etal-2023-explore}. Existing approaches, shown in Figure~\ref{fig:framework}, mainly fall into two main categories. The first category uses domain-specific corpora~\cite{lewkowycz2022solving,taylor2022galactica,bonito:aclfindings24}, but these resources are often proprietary, expensive, or simply unavailable. The second category requires human experts to write seed examples for in-context learning (ICL) prompts~\cite{peng2023instruction,zhang2023targen,wang-etal-2023-self-instruct,wan-etal-2023-explore}. Both approaches demand significant domain expertise, which limits their scalability and hinders efficient domain adaptation.

This paper introduces \textsc{DS\textsuperscript{2}-Instruct} (\textbf{\underline{D}}omain-\textbf{\underline{S}}pecific \textbf{\underline{D}}ata \textbf{\underline{S}}ynthesis), a zero-shot framework that systematically generates high-quality, domain-specific instruction data. Our framework requires no human supervision, seed examples, or specialized corpora. \textsc{DS\textsuperscript{2}-Instruct} addresses three core challenges in domain-specific data synthesis:

\begin{itemize}[leftmargin=*,itemsep=-1pt,topsep=-0.8pt,label={}]
\item \ding{182} \textit{How to systematically capture comprehensive domain-specific knowledge?} To ensure \textit{knowledge coverage}, it uses task-informed keyword generation with bi-directional expansion and retrieval augmentation to discover domain concepts from foundational to specialized terminology. 
\item \ding{183} \textit{How to generate cognitively diverse questions?} To promote \textit{instruction diversity}, it applies Bloom's Taxonomy~\cite{bloom1956taxonomy,forehand2010bloom} to create instructions across six cognitive levels. 
\item \ding{184} \textit{How can we ensure the generated instruction-response pairs are of high quality?} To guarantee \textit{data quality}, it employs self-consistency filtering~\cite{wang2023selfconsistency} to identify and remove flawed instruction-response pairs.
\end{itemize}
We apply \textsc{DS\textsuperscript{2}-Instruct} to generate instruction datasets across diverse domains, including mathematics, finance, and biomedical sciences, using \texttt{Qwen2.5-72B-Instruct}~\cite{qwen2} as the generator. We then fine-tune several models, including \texttt{Llama3-8B}~\cite{llama3modelcard}, \texttt{Mistral-7B}~\cite{jiang2023mistral}, and \texttt{Qwen2.5-7B}~\cite{qwen2}, on our generated data. Evaluations on seven domain-specific benchmarks show consistent improvements over the base models. Our contributions are threefold:

\begin{itemize}
[leftmargin=*,itemsep=0.2pt,topsep=0.5pt]
\item We introduce a zero-shot framework that generates domain-specific instruction data without requiring human supervision, seed examples, or specialized corpora.
\item We show that incorporating cognitive diversity from Bloom's Taxonomy improves a model's reasoning performance.
\item We demonstrate through comprehensive evaluation that models trained on our data substantially outperform those trained with existing methods.
\end{itemize}

\section{Related Work}
\label{sec:preliminary_related}
\noindent\textbf{General-Purpose Instruction Data Synthesis.}
The emergence of instruction-following capabilities in LLMs has been driven by advances in instruction tuning methodologies~\cite{wei2021finetuned, sanh2021multitask, ouyang2022training}. Methods for synthesizing general-purpose instruction data have evolved through several paradigms. Early approaches use manually curated datasets to tune models~\cite{mishra2021cross, wei2021finetuned, sanh2021multitask, wang-etal-2022-super}. This manual process is expensive and does not scale. A second paradigm uses LLMs to expand a small set of human-written seeds~\cite{wang-etal-2023-self-instruct, taori2023stanford}. Recent work has explored more systematic approaches to data synthesis. Some methods employ keyword-based or topic-guided generation to enhance diversity and coverage~\cite{ding-etal-2023-enhancing, mukherjee2023orca}. A third paradigm employs LLMs to convert unannotated web documents into instruction-tuning datasets~\cite{wang-etal-2022-super, jiang-etal-2025-instruction}. 

\noindent\textbf{Domain-Specific Instruction Tuning.}
While general-purpose instruction tuning methods have shown success, there is increasing demand for domain-specific instruction tuning that remains underexplored. The primary difficulty is acquiring specialized knowledge. One common strategy converts existing domain-specific documents into instruction data~\cite{bonito:aclfindings24,singhal2023med,katz2023gpt,yang2023fingpt}. However, such documents are often proprietary or difficult to obtain. Other methods avoid the need for document collections but instead require human-written seed data or are confined to narrow NLP tasks~\cite{zhang2023targen, wan-etal-2023-explore}. Concurrently, \citet{guo2025synthetic} leverage explicit task definitions to generate synthetic data for reinforcement learning fine-tuning, while reusing an initial set of 500 seed samples. In contrast, our \textsc{DS\textsuperscript{2}-Instruct} framework systematically generates domain-specific data without requiring documents, seeds, or human supervision.

\section{Method}
\label{sec:methods}
In this section, we present \textsc{DS\textsuperscript{2}-Instruct}, a systematic framework for generating high-quality task-specific instruction-tuning datasets that enable effective adaptation of LLMs to specialized tasks. Our approach addresses the critical challenge of creating diverse and task-relevant training data without requiring extensive human annotation. 

\subsection{Problem Formulation}
\begin{figure*}[t]
   \centering
   \includegraphics[width=1\textwidth]{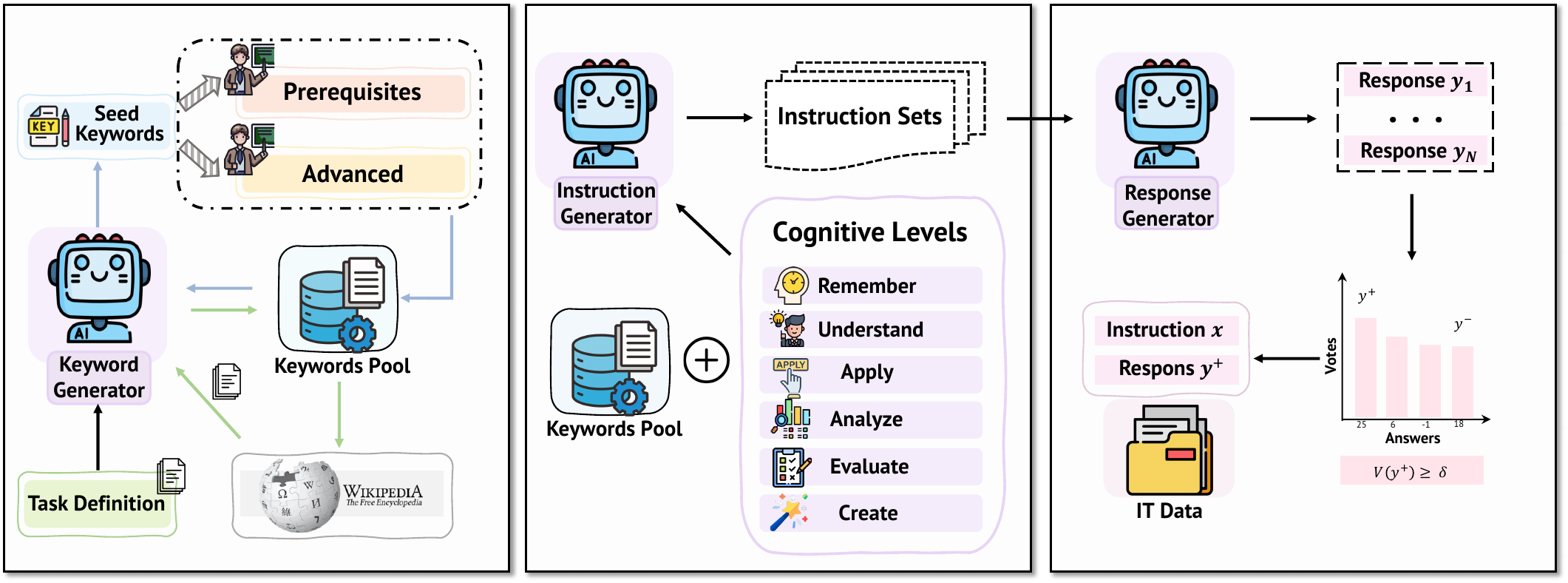}
   \caption{Overview of the \textsc{DS\textsuperscript{2}-Instruct} framework. Given only a task definition, our approach systematically generates domain-specific instruction-tuning datasets through three stages: \ding{182} keyword generation using bi-directional expansion and retrieval augmentation, \ding{183} cognitive-level instruction generation based on Bloom's Taxonomy, and \ding{184} self-consistency filtering for quality assurance.}
   \label{fig:framework_overview}
   \vspace{-15pt}
\end{figure*}
Let $T_{desc}$ be a unified task description that encompasses both domain knowledge requirements and task format specifications. Our objective is to generate a high-quality, task-aligned instruction dataset $\mathcal{S} = \{(x_i, y_i)\}_{i=1}^N$, where $x_i$ represents an instruction and $y_i$ represents the corresponding response. 

\subsection{Task-Informed Keyword Generation}
\label{subsec:keyword_generation}
The keyword generation phase constructs a task-specific knowledge base by systematically identifying and expanding a set of domain-specific concepts. This process begins with a small seed set of keywords derived directly from the task description and iteratively expands it through bi-directional
expansion and retrieval augmentation to ensure comprehensive coverage of the relevant knowledge areas while strictly maintaining relevance to the target task.

\noindent\textbf{Initial Keyword Seeding.}
We initiate the process by generating a foundational set of keywords. Given the task description $T_{desc}$, we prompt an LLM $\mathcal{M}$ to produce an initial set of $n$ keywords that are most representative of the core domain:
\begin{equation}
K_0 = \text{Parse}(\mathcal{M}(\text{Prompt}_i, T_{desc}, n))
\label{eq:initial_keywords}
\end{equation}
where $\text{Prompt}_i$ represents the initial prompt used for keyword generation, and $\text{Parse}(\cdot)$ is a function that extracts a structured list of keywords from the raw textual output of the LLM. This initial set $K_0$ serves as the foundational knowledge base for subsequent retrieval-augmented expansion.

\noindent\textbf{Bi-Directional Keyword Expansion.}
To ensure comprehensive coverage of the knowledge space for a specific domain, we propose a bi-directional expansion strategy. Given the initial keyword set $K_0$, we iteratively use the LLM to systematically generate both prerequisite foundational concepts and advanced specialized topics, enabling full exploration of the domain's knowledge hierarchy.

At each iteration, we randomly sample a subset of keywords from the current pool and use them as in-context learning examples to guide the expansion process. The prerequisite expansion identifies fundamental concepts, basic terminology, and foundational principles that learners must understand before engaging with the sampled concepts. Conversely, the advanced expansion explores specialized subfields, cutting-edge developments, and expert-level topics that build upon the current knowledge base. After each iteration, we integrate the newly generated concepts into the expanding keyword pool. The random sampling ensures that each expansion explores varied conceptual relationships and prevents the generation from converging prematurely on a limited subset of domain knowledge. This systematic and iterative exploration of both foundational and advanced concepts enables comprehensive domain coverage, ensuring that the resulting instruction dataset addresses diverse concepts within the target domain.

\vspace{-5pt}
\begin{tcolorbox}[
    colback=white,
    colframe=promptboxblue,
    colbacktitle=promptboxblue,
    coltitle=white,
    fonttitle=\small\bfseries,
    fontupper=\small,
    title=Prompt Template for Bi-Directional Keyword Expansion,
    breakable,
    arc=1mm,
    boxrule=1.2pt
]
\textbf{Task Context}: You are an expert in the domain related to: {\color{red!70!black}[Task description $T_{desc}$]}.

\textbf{Sample Keywords}: {\color{red!70!black}[Randomly sampled keywords from current pool]}.

\textbf{Instructions}: Based on the sample keywords, generate new concepts in two directions:

\begin{enumerate}
[leftmargin=*,itemsep=0.2pt,topsep=1pt]
    \item {\color{purple!70!black}\textbf{Prerequisite Concepts}}: What fundamental concepts, basic terminology, or foundational principles should learners understand BEFORE studying the sample keywords?
    \item {\color{purple!70!black}\textbf{Advanced Concepts}}: What specialized subfields, cutting-edge developments, or expert-level topics BUILD UPON the sample keywords?
\end{enumerate}
\end{tcolorbox}

\noindent\textbf{Retrieval-Augmented Keyword Extraction.}
After obtaining the expanded keyword set from bi-directional expansion, we enhance our knowledge base through corpus-based retrieval to reduce hallucination risks and ensure deeper domain coverage~\cite{biderman2022datasheet,izacard2023atlas,borgeaud2022improving,lewis2020retrieval}. We construct a diverse retrieval corpus by sampling documents from the Pile~\cite{gao2020pile}, a multidomain collection of high-quality human-written documents. Specifically, we sample from five complementary datasets: ArXiv, FreeLaw, StackExchange, Wikipedia, and Github. This multidomain approach ensures comprehensive coverage across different writing styles, technical depths, and domain-specific terminology.

For retrieval, given the task description $T_{desc}$ and our expanded keyword set, we randomly sample keyword subsets and combine them with the task description to form retrieval queries. For each query, the retriever scores all documents in our corpus based on term frequency, inverse document frequency, and document length normalization, returning the top-$k$ most relevant passages~\citep{robertson2009probabilistic}. These retrieved passages provide authoritative, grounded context about related concepts and domain-specific knowledge. Given the retrieved passages and the current keyword list, we prompt the LLM to extract additional keywords directly from the passage content. 

\begin{tcolorbox}[
    colback=white,
    colframe=promptboxblue,
    colbacktitle=promptboxblue,
    coltitle=white,
    fonttitle=\small\bfseries,
    fontupper=\small,
    title=Prompt Template for Retrieval-Augmented Keyword Extraction,
    breakable,
    arc=1mm,
    boxrule=1.2pt
]
\textbf{Task Context}: You are an expert in the domain related to: {\color{red!70!black}[Task description $T_{desc}$]}.

\textbf{Current Keywords}: {\color{red!70!black}[List of expanded keywords]}.

\textbf{Retrieved Passages}: The following passages contain comprehensive domain knowledge:
{\color{blue!70!black}[Top-k retrieved passages]}

\textbf{Instructions}: Extract additional domain-specific keywords directly from the retrieved passages that are missing from the current list.
\end{tcolorbox}

\subsection{Instruction Generation Across Cognitive Levels}
\label{subsec:instruction_generation}
Traditional instruction generation techniques often produce instructions that test only surface-level recall or follow repetitive patterns, failing to assess deeper cognitive skills essential for true domain mastery. Drawing inspiration from educational assessment research, we develop a cognitive-task integration framework based on Bloom's Taxonomy~\cite{bloom1956taxonomy, forehand2010bloom} that systematically generates instructions across six cognitive levels, from basic recall to creative synthesis, ensuring diverse cognitive complexity while maintaining strict task format compliance.

\noindent\textbf{Cognitive Framework Integration.} We systematically integrate Bloom's Taxonomy~\cite{bloom1956taxonomy,forehand2010bloom} to ensure comprehensive cognitive coverage in our instruction generation. The taxonomy consists of six hierarchical cognitive levels $\mathcal{B} = \{R, U, Ap, An, E, C\}$ corresponding to \textit{Remember}, \textit{Understand}, \textit{Apply}, \textit{Analyze}, \textit{Evaluate}, and \textit{Create}. By explicitly incorporating each level into our generation process, we ensure that the resulting instructions evaluate diverse cognitive skills ranging from basic knowledge retention to complex creative synthesis:

\begin{tcolorbox}[
  colback=white,
  colframe=promptboxblue,
  colbacktitle=promptboxblue,
  coltitle=white,
  fonttitle=\small\bfseries,
  fontupper=\small,
  title=Cognitive Levels for Instruction Generation,
  breakable,
  arc=1mm,
  boxrule=1.2pt
]
\textcolor{red!70!black}{\textbf{Remembering}}: Tests factual recall, definitions, and basic terminology.

\textcolor{orange!70!black}{\textbf{Understanding}}: Requires conceptual explanations and interpretations.

\textcolor{green!70!black}{\textbf{Applying}}: Demands practical application in concrete scenarios.

\textcolor{blue!70!black}{\textbf{Analyzing}}: Involves decomposing problems and identifying patterns.

\textcolor{purple!70!black}{\textbf{Evaluating}}: Requires critical assessment and justified decisions.

\textcolor{brown!70!black}{\textbf{Creating}}: Necessitates synthesizing information into novel solutions.
\end{tcolorbox}

\noindent\textbf{Keyword Matching Strategies.} To maximize instruction diversity, we combine different question types with two complementary keyword approaches. Single-keyword instructions ensure comprehensive coverage of individual concepts across all cognitive levels, while paired-keyword instructions explore relationships between concepts and increase overall diversity through multi-concept integration. For single keywords, we pair each keyword $k_j \in K^*$ with each cognitive level $b \in \mathcal{B}$. For keyword pairs $(k_i, k_j)$, we focus on cognitive levels $\mathcal{B}_{rel} = \{U, Ap, An, E\}$ that naturally accommodate relational reasoning. This generates a comprehensive instruction set covering both individual concept mastery and inter-concept relationships.

\begin{tcolorbox}[
 colback=white,
 colframe=promptboxblue,
 colbacktitle=promptboxblue,
 coltitle=white,
 fonttitle=\small\bfseries,
 fontupper=\small,
 title=Prompt Template for Instruction Generation,
 breakable,
 arc=1mm,
 boxrule=1.2pt
]
\textbf{Single-Keyword Instructions}: Given keyword {\color{red!70!black}[$k_j$]} and cognitive level {\color{blue!70!black}[$b$]}, generate a high-quality instruction following task requirements {\color{red!70!black}[$T_{desc}$]}. Ensure the keyword is the central focus and use appropriate domain terminology.

\textbf{Paired-Keyword Instructions}: Given keyword pair {\color{red!70!black}[$k_i$, $k_j$]} and cognitive level {\color{blue!70!black}[$b$]}, generate an instruction that explores relationships between both keywords while following task requirements {\color{red!70!black}[$T_{desc}$]}. Focus on multi-concept integration and comparative reasoning.
\end{tcolorbox}

\subsection{Instruction Filtering and Response Generation}
\label{subsec:response_generation}
After generating a large pool of candidate instructions, we employ a systematic filtering and response generation process to ensure high-quality instruction-response pairs. This process consists of two key stages: self-consistency filtering to remove low-quality instructions and final response generation using majority voting.

\noindent\textbf{Self-Consistency Filtering.} To identify and remove low-quality instructions where the model exhibits uncertainty, we employ a self-consistency mechanism, which has been used for improving performance on reasoning tasks by generating multiple solutions using LLMs and choosing the most frequent final answer~\cite{wang2023selfconsistency, yu2025cot, shi-etal-2022-natural, li2022competition, prasad2024self}. For each candidate instruction $x \in \mathcal{I}$, we prompt the LLM $\mathcal{M}$ to generate $N$ independent responses $\{y_1, y_2, \ldots, y_N\}$. We then extract the final answer from each response using an answer extraction function $\text{ans}(\cdot)$ and compute the vote function $V(\cdot)$:
\vspace{-10pt}
\begin{equation}
V(y) = \frac{1}{N} \sum_{m=1}^{N} \mathbf{1}(\text{ans}(y_m) = \text{ans}(y))
\label{eq:vote_function}
\end{equation}
The vote function $V(\cdot)$ returns the relative frequency of each final answer, where $\mathbf{1}(\cdot)$ is the indicator function. This approach quantifies instruction quality based on response consistency. The intuition is that it is less likely to consistently generate the same incorrect answer across multiple attempts. Therefore, instructions that consistently produce the same answer across multiple generations are considered high-quality, while those yielding diverse or conflicting responses indicate ambiguity or poor formulation. We filter out instructions where $\arg\max_{y \in \{y_1, \ldots, y_N\}} V(y) < \tau$, retaining only those with sufficient response consistency.

\noindent\textbf{Final Response Generation.} For each high-quality instruction $x_i \in \mathcal{I}_{filtered}$, we generate the final response $y_i$ by selecting the answer that achieved the highest vote frequency during the consistency evaluation. The final dataset $\mathcal{S} = \{(x_i, y_i)\}_{i=1}^{|\mathcal{I}_{filtered}|}$ combines instructions generated from both single-keyword and paired-keyword strategies, providing comprehensive cognitive coverage while maintaining high quality through self-consistency validation.

\section{Experiments}
\label{sec:experiments}

\subsection{Experimental Setup}
\noindent\textbf{Datasets and Benchmark Tasks.} To evaluate the ability of \textsc{DS\textsuperscript{2}-Instruct} to adapt to specific domains, we use seven different publicly available datasets spanning the challenging domains of mathematics, medicine, logical reasoning, science, and finance~\cite{xie2023pixiu,jin2021disease,liu2020logiqa,hendrycks2021measuring,cobbe2021gsm8k,jin-etal-2019-pubmedqa,rein2024gpqa}. These benchmarks were chosen to test a wide array of reasoning and knowledge-intensive skills. A detailed summary of the datasets, their respective task types, sizes, and primary evaluation metrics is provided in Appendix \ref{sec:dataset}.\\
\noindent\textbf{Baseline Methods.} We compare \textsc{DS\textsuperscript{2}-Instruct} against four baseline approaches: (1) \textit{Zero-Shot} evaluation using vanilla pre-trained models without fine-tuning, (2) a domain-aware version of \textit{Self-Instruct}~\cite{wang-etal-2023-self-instruct} where we use human-crafted domain-specific seed questions and iteratively expand the instruction collection by randomly selecting five examples during each iteration, (3) \textit{InstructMix}~\cite{xu2023data} which first extracts core skills using LLM, then generates instruction-response pairs that exhibit randomly chosen combinations of these skills, and (4) \textit{ExploreInstruct}~\cite{wan-etal-2023-explore} that leverages the exploration power of LLMs to actively navigate the domain space and obtain domain-focused instructions. \\
\noindent\textbf{Base Models.} For the main experiment, we use \texttt{Qwen2.5-72B-Instruct}~\cite{qwen2} as the data generator for all baseline methods and \textsc{DS\textsuperscript{2}-Instruct} to ensure fair comparison across different instruction generation approaches. For fine-tuning evaluation, we employ three target models: \texttt{Qwen2.5-7B}~\cite{qwen2}, \texttt{Llama-3.1-8B}~\cite{touvron2023llama}, and \texttt{Mistral-7B-v0.1}~\cite{jiang2023mistral}. This configuration allows us to assess the transferability of generated instructions across different model architectures while maintaining consistency in the data generation process.\\
\begin{table*}[t]
\centering
\caption{Statistics of generated instruction datasets across different domains.}
\scalebox{0.87}{
   \begin{tabular}{lcccccccc}
   \toprule
   \textbf{Metric} & \textbf{CFA} & \textbf{PubMedQA} & \textbf{MedQA} & \textbf{GPQA} & \textbf{LogiQA} & \textbf{GSM8K} & \textbf{MATH} \\
   \midrule
   Avg. Instruction Length (words) $\uparrow$ & 154.92 & 119.73 & 140.65 & 144.98 & 169.39 & 108.49 & 116.24 \\
   Unique Verb-Noun Pairs $\uparrow$ & 5,706 & 2,572 & 6,834 & 6,684 & 5,211 & 1,855 & 1,630 \\
   Avg. Pair Occurrences $\downarrow$ & 2.69 & 2.32 & 2.16 & 2.02 & 2.33 & 2.25 & 6.19 \\
   Std. Dev. of Occurrences $\downarrow$ & 17.01 & 5.35 & 7.54 & 6.27 & 18.94 & 9.94 & 148.77 \\
   \bottomrule
   \label{tab:data}
   \vspace{-10pt}
   \end{tabular}
}
\label{tab:data_statistics}
\end{table*}
\noindent\textbf{Implementation Details.} 
To ensure fair comparison across all methods, we utilize 6000 instruction-response pairs for each baseline and task. For our proposed \textsc{DS\textsuperscript{2}-Instruct}, we begin with 50 initial seed keywords and perform bi-directional expansion where we generate 5 new keywords in each direction per iteration, conducting 100 iterations total. We employ LoRA~\cite{hu2021lora} for parameter-efficient fine-tuning with rank $r=8$ and $\alpha = 16$ for parameter-efficient adaptation. For detailed implementation specifications, hyperparameter configurations, and training details for both settings, please refer to Appendix~\ref{sec:implementation}.

\subsection{Data-Centric Analysis}
\textbf{Data Statistics.} Table~\ref{tab:data} presents essential statistics of the data generated by our proposed \textsc{DS\textsuperscript{2}-Instruct} framework. We analyze comprehensive metrics including average instruction length, linguistic diversity measures such as unique verb-noun pairs, average occurrence of verb-noun pairs in each instruction, and standard deviation of average occurrences across the dataset. The linguistic diversity metrics are particularly important as they indicate the richness of the generated instructions. Higher numbers of unique verb-noun pairs suggest more varied linguistic patterns, while lower average occurrences indicate better distribution without excessive repetition~\cite{wei2021finetuned, sanh2021multitask}. The standard deviation of occurrences measures the consistency of term distribution, where lower values generally indicate more balanced vocabulary usage~\cite{pmlr-v202-longpre23a}.
\begin{figure}[h!]
    \centering
    \begin{subfigure}{\columnwidth}
        \centering
        \includegraphics[width=0.7\columnwidth]{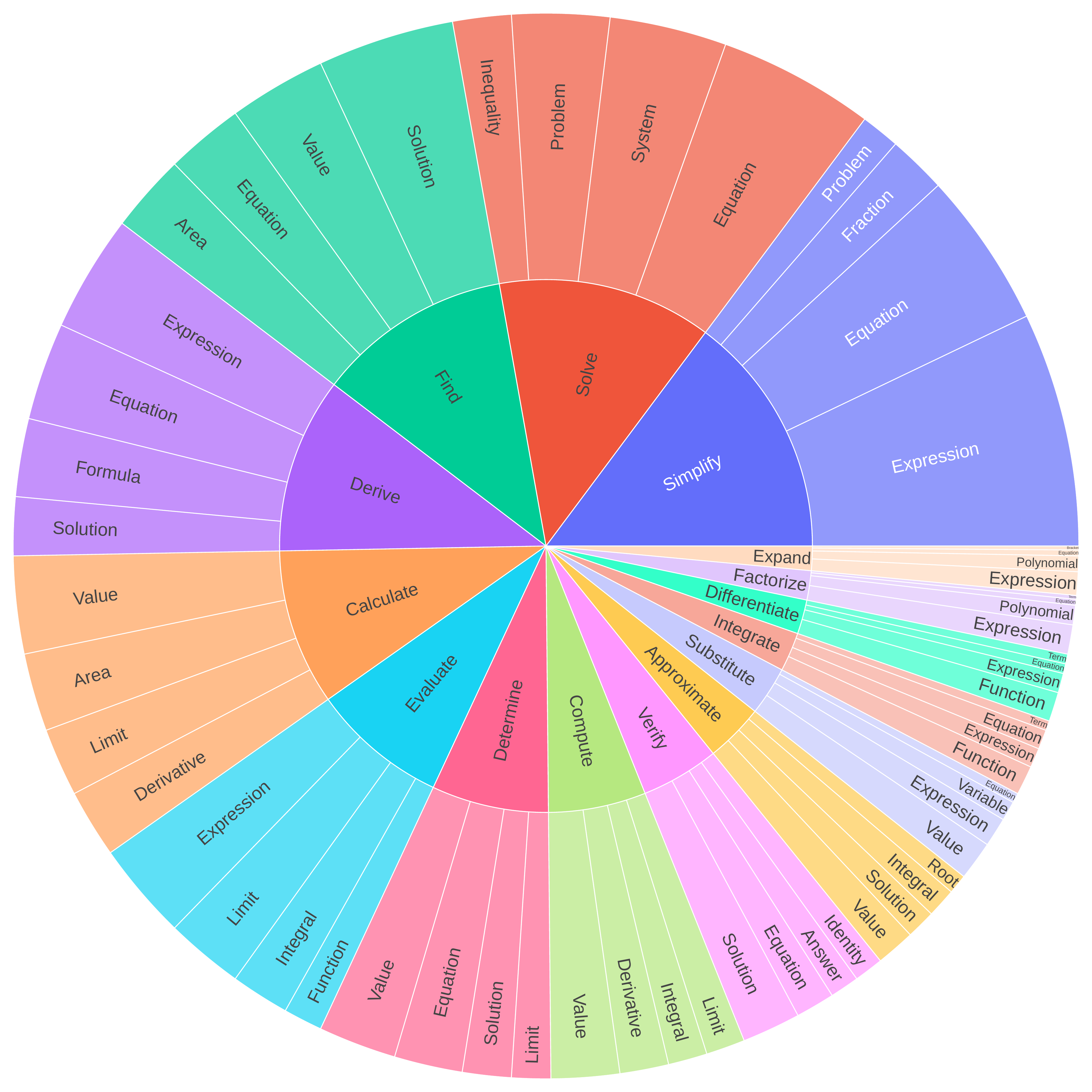}
        
        {(a) Domain-Aware Self-Instruct}
        \label{fig:verb_noun_selfinstruct}
    \end{subfigure}
    \begin{subfigure}{\columnwidth}
        \centering
        \includegraphics[width=0.7\columnwidth]{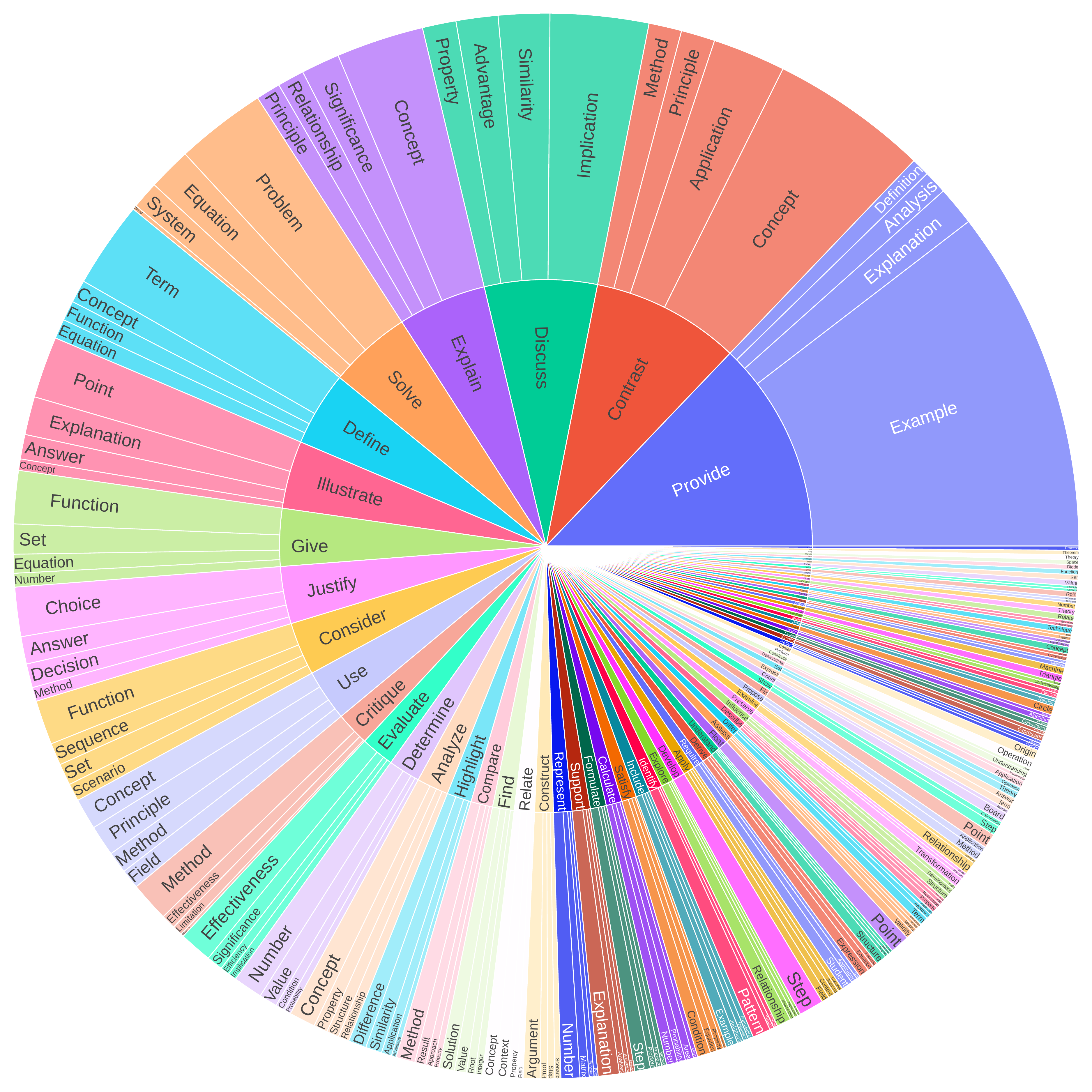}
        
        {(b) \textsc{DS\textsuperscript{2}-Instruct}}
        \label{fig:verb_noun_domaininstruct}
    \end{subfigure}
    \caption{Root verb-noun pairs in instructions of (a) Domain-Aware Self-Instruct and (b) \textsc{DS\textsuperscript{2}-Instruct} in the math domain. The inner circle represents root verbs and the outer circle represents direct nouns. For each verb, we show the top 4 verb-noun pairs.}
    \vspace{-20pt}
    \label{fig:verb_noun_diversity}
\end{figure}

\begin{table*}[t]
\centering
\caption{
Performance comparison across domain-specific benchmarks. Results show accuracy (\%) for multiple-choice tasks and exact match (\%) for problem-solving tasks. For each model family, the best results are \textbf{bold} and second-best are \underline{underlined}.
}
\scalebox{0.9}{
\begin{tabular}{c lccccccc c}
\toprule
&  \multirow{2}{*}{\textbf{Methods}} & \textbf{Finance} & \multicolumn{2}{c}{\textbf{Medicine}}  & \textbf{Science} & \textbf{Logic} & \multicolumn{2}{c}{\textbf{Math}} & \multirow{2}{*}{\textbf{Avg.}} \\ 
\cmidrule(lr){3-3} \cmidrule(lr){4-5} \cmidrule(lr){6-6} \cmidrule(lr){7-7} \cmidrule(lr){8-9}
&   & CFA & PubMedQA & MedQA & GPQA & LogiQA & GSM8K & MATH &    \\ 
\midrule
\multirow{5}{*}{\rotatebox[origin=c]{90}{\texttt{Qwen2.5}}} 
& Zero-Shot & 53.71 & 60.34 & 48.35 & 26.32 & 32.77 & 71.53 & 49.44 & 48.92 \\
\cmidrule{2-10}
& Self-Instruct & 55.83 & 61.47 & \underline{48.92} & \textbf{27.18} & 42.56 & 72.38 & 53.67 & 51.72 \\
& InstructMix & \underline{57.91} & \underline{62.15} & 47.83 & 26.09 & \underline{43.88} & 75.42 & \underline{56.24} & \underline{53.07} \\
& ExploreInstruct & 56.47 & 60.89 & 47.15 & 25.93 & 41.19 & \underline{76.67} & 51.38 & 51.10 \\
\rowcolor[HTML]{EAF3FC} & \textsc{DS\textsuperscript{2}-Instruct} & \textbf{58.34} & \textbf{63.62} & \textbf{50.98} & \underline{26.35} & \textbf{44.29} & \textbf{78.94} & \textbf{60.12} & \textbf{54.95} \\
\midrule
\multirow{5}{*}{\rotatebox[origin=c]{90}{\texttt{Llama3}}}
& Zero-Shot & 9.76 & 24.92 & 27.38 & 12.51 & 9.77 & 12.60 & 3.45 & 14.34 \\
\cmidrule{2-10}
& Self-Instruct & 33.72 & 25.34 & 31.58 & \underline{19.87} & \underline{24.16} & 19.83 & 11.47 & 23.71 \\
& InstructMix & \underline{36.89} & \underline{27.41} & \underline{33.29} & 17.52 & 15.88 & \underline{35.64} & \underline{14.92} & \underline{25.94} \\
& ExploreInstruct & 34.58 & 24.77 & 30.46 & 18.24 & 22.35 & 33.81 & 13.06 & 25.32 \\ 
\rowcolor[HTML]{EAF3FC} & \textsc{DS\textsuperscript{2}-Instruct} & \textbf{56.94} & \textbf{48.56} & \textbf{39.15} & \textbf{30.35} & \textbf{35.33} & \textbf{58.34} & \textbf{23.16} & \textbf{42.83} \\
\midrule
\multirow{5}{*}{\rotatebox[origin=c]{90}{\texttt{Mistral}}}
& Zero-Shot & 4.61 & 14.92 & 14.03 & 8.64 & 9.52 & 14.75 & 1.42 & 9.70 \\
\cmidrule{2-10}
& Self-Instruct & 14.26 & 26.83 & \underline{39.74} & 8.15 & \underline{29.67} & 18.52 & 5.94 & 20.30 \\
& InstructMix & \underline{42.15} & \underline{31.48} & 30.15 & \underline{23.91} & 26.72 & \underline{38.26} & 8.67 & \underline{30.29} \\
& ExploreInstruct & 38.64 & 28.35 & 37.92 & 19.47 & 25.18 & 17.93 & \underline{11.74} & 24.86 \\
\rowcolor[HTML]{EAF3FC} & \textsc{DS\textsuperscript{2}-Instruct} & \textbf{55.72} & \textbf{39.11} & \textbf{42.99} & \textbf{30.16} & \textbf{32.23} & \textbf{42.75} & \textbf{21.82} & \textbf{37.83} \\
\bottomrule
\end{tabular}
}
\label{tab:main_results}
\end{table*}

\noindent\textbf{Diversity.} To evaluate instruction diversity, we analyze verb-noun pairs with frequency above 10 in the mathematics domain, comparing Domain-Aware Self-Instruct and \textsc{DS\textsuperscript{2}-Instruct} in Figure~\ref{fig:verb_noun_diversity}. The sunburst plots show root verbs (inner circle) and direct nouns (outer circle), with top 4 pairs displayed for each verb. Figure~\ref{fig:verb_noun_diversity}(a) reveals that Domain-Aware Self-Instruct produces highly concentrated distributions with significantly fewer unique verb-noun pairs, clustering around a limited set of computational verbs like ``Calculate'', ``Solve'' and ``Find'' paired with basic objects such as ``Expression'' and ``Equation''. This concentration indicates reduced diversity and potential overfitting to common instruction patterns. In contrast, Figure~\ref{fig:verb_noun_diversity}(b) demonstrates that \textsc{DS\textsuperscript{2}-Instruct} generates substantially more unique verb-noun pairs with more uniformly distributed patterns and diverse mathematical concepts.

\begin{table}[hbtp]
\centering
\caption{Quality assessment of generated instruction-response pairs.}
\label{tab:quality}
\scalebox{0.88}{
\begin{tabular}{lcc}
\toprule
\textbf{Quality Metric} & \textbf{GSM8K (\%)} & \textbf{MedQA (\%)} \\
\midrule
Valid instruction & 96.83 & 92.42 \\
Valid response & 91.26 & 94.89 \\
Domain-appropriate & 94.91 & 90.87 \\
\bottomrule
\vspace{-30pt}
\end{tabular}
}
\end{table}
\noindent\textbf{Quality.} In addition to statistics and diversity of generated data, we evaluate the quality of the generated instructions and responses. Following previous work~\cite{wang-etal-2023-self-instruct}, we randomly sample 500 instruction-response pairs and ask an expert annotator (an author of this work) to label whether each instance is valid in terms of instruction clarity, response correctness, and domain appropriateness. Table~\ref{tab:quality} reports the percentage of valid instructions and responses across datasets. We also provide a detailed analysis of Bloom's Taxonomy to quantify the proportion of generated instructions at each cognitive level and present illustrative case studies in Appendix~\ref{sec:casestudy}.
\subsection{Main Results}
We evaluate \textsc{DS\textsuperscript{2}-Instruct} against several representative instruction-generation baselines across seven domain-specific benchmarks and three model families: \texttt{Qwen2.5}, \texttt{Llama3}, and \texttt{Mistral}. As shown in Table~\ref{tab:main_results}, \textsc{DS\textsuperscript{2}-Instruct} consistently achieves the best average performance within each model family, outperforming all baseline methods across nearly all domains. 

Notably, the performance gains are most pronounced for weaker base models. For \texttt{Mistral}, which exhibits the lowest zero-shot performance, \textsc{DS\textsuperscript{2}-Instruct} improves the average score to 37.83\%. Similarly, on \texttt{Llama3}, our method raises the average performance from 14.34\% in the zero-shot setting to 42.83\%, substantially exceeding prior instruction-generation approaches. Even for the stronger \texttt{Qwen2.5} model, \textsc{DS\textsuperscript{2}-Instruct} delivers consistent improvements. The consistent improvements across diverse domains and model architectures highlight the broad applicability and effectiveness of our framework. 

\subsection{Additional Experiments}
To further validate the robustness and generalizability of our approach, we conduct additional experiments examining two critical dimensions: the impact of training data scale and the effectiveness across models of varying sizes. 

\begin{figure}[!ht]
\centering
\vspace{-10pt}
\includegraphics[width=0.38\textwidth]{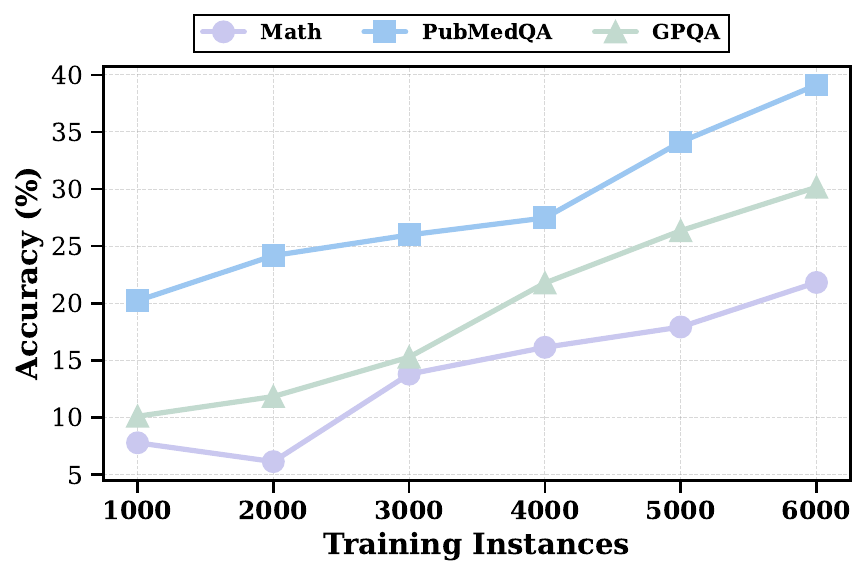}
\caption{Impact of Training Data Size on Model Performance.}
\label{fig:training_scale}
\end{figure}

\noindent\textbf{Impact of Training Data Scale.}
We investigate how training data size affects model performance by conducting experiments with varying numbers of training instances ranging from 1,000 to 6,000 samples using \texttt{Mistral-7B} as the base model. As illustrated in Figure~\ref{fig:training_scale}, all three datasets demonstrate consistent performance improvements as training data increases, though with distinct scaling patterns. Notably, even at 6,000 samples, the curves have not fully saturated, suggesting that further performance improvements may be achievable with larger training datasets. This scaling behavior validates the effectiveness of our data synthesis approach and indicates that the quality of synthesized instructions remains high even as quantity increases.

\begin{figure}[!ht]
\centering
\vspace{-10pt}
\includegraphics[width=\linewidth]{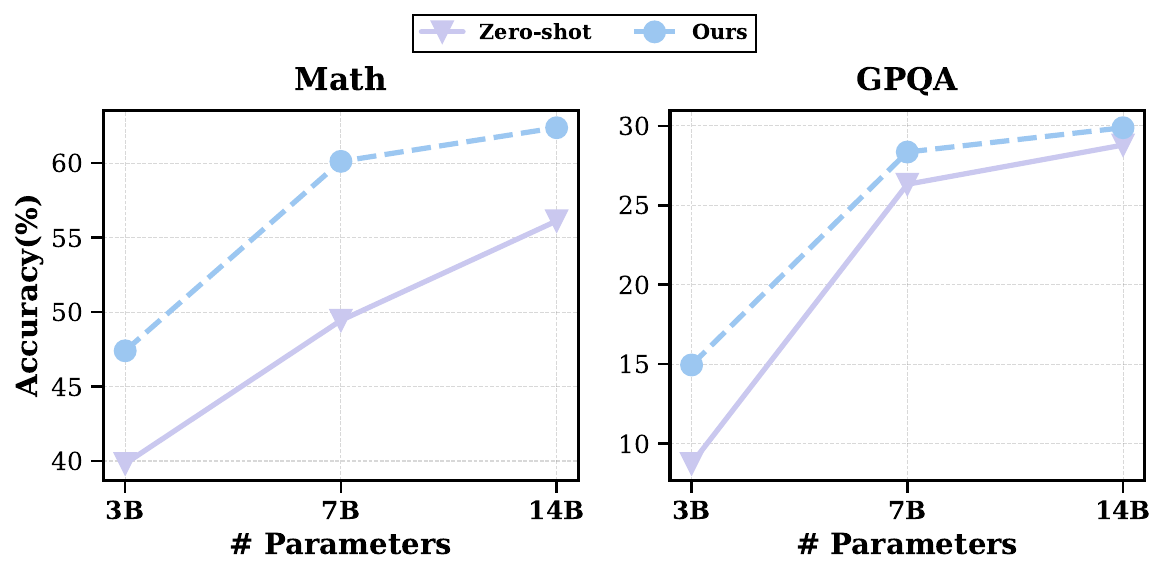}
\caption{Performance on Models of Different Sizes. We evaluate the Qwen model family on Math and GPQA datasets.}
\vspace{-10pt}
\label{fig:model_scaling}
\end{figure}

\noindent\textbf{Performance Across Model Scales.}
To assess the scalability of our approach, we evaluate models of different sizes from the \texttt{Qwen} family (3B, 7B, 14B parameters) on Math and GPQA benchmarks. Figure~\ref{fig:model_scaling} shows that our \frameworkname\xspace method consistently outperforms zero-shot baselines across all model scales. Notably, the performance gains diminish as model size increases, suggesting that larger models may rely less heavily on domain-specific instruction data and can better leverage their inherent reasoning capabilities.

\subsection{Ablation Study}
To comprehensively evaluate the effectiveness of our \textsc{DS\textsuperscript{2}-Instruct} framework, we conduct systematic ablation studies analyzing three core components: (1) the keyword generation mechanism with bi-directional expansion and retrieval augmentation, (2) the cognitive level integration based on Bloom's Taxonomy, and (3) the self-consistency filtering approach. These studies are performed across four diverse datasets using \texttt{Mistral-7B} to understand each component's individual contribution to overall performance. 

\begin{figure}[htbp]
    \centering
    \vspace{-20pt}
    \includegraphics[width=0.48\textwidth]{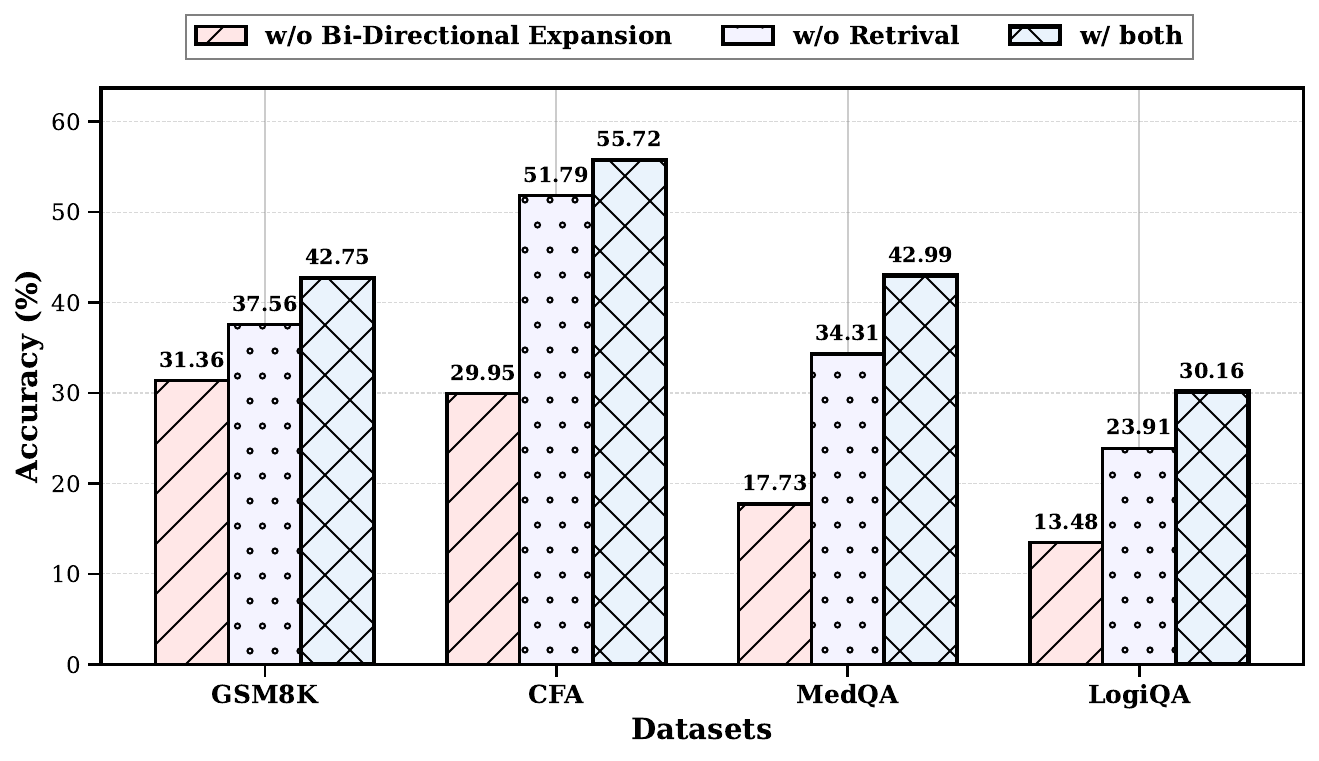}
    \caption{Ablation study evaluating the effectiveness of keyword generation components.}
    \vspace{-15pt}
    \label{fig:ablation_keywords}
\end{figure}

\noindent\textbf{Effectiveness of Keywords Generation.} To validate our keyword generation components, we conduct ablation studies across four datasets using \texttt{Mistral-7B}. We compare three configurations: (1) \textit{w/o Bi-Directional Expansion}, which uses direct prompting to obtain seed keywords without iterative expansion, (2) \textit{w/ Retrieval}, which expands keywords through LLM generation without Wikipedia-based retrieval, and (3) \textit{w/ Both}, our complete framework with retrieval-augmented expansion. Figure~\ref{fig:ablation_keywords} shows the critical importance of both components. Without bi-directional expansion, performance is limited across all datasets. Adding LLM-based expansion provides moderate improvements. However, incorporating retrieval-augmented expansion yields substantial gains, with the full framework outperforming the baseline. The synergistic effect of combining systematic keyword expansion with authoritative Wikipedia sources validates our integrated approach for comprehensive domain coverage.

\begin{figure}[!ht]
    \centering
    \vspace{-10pt}\includegraphics[width=0.48\textwidth]{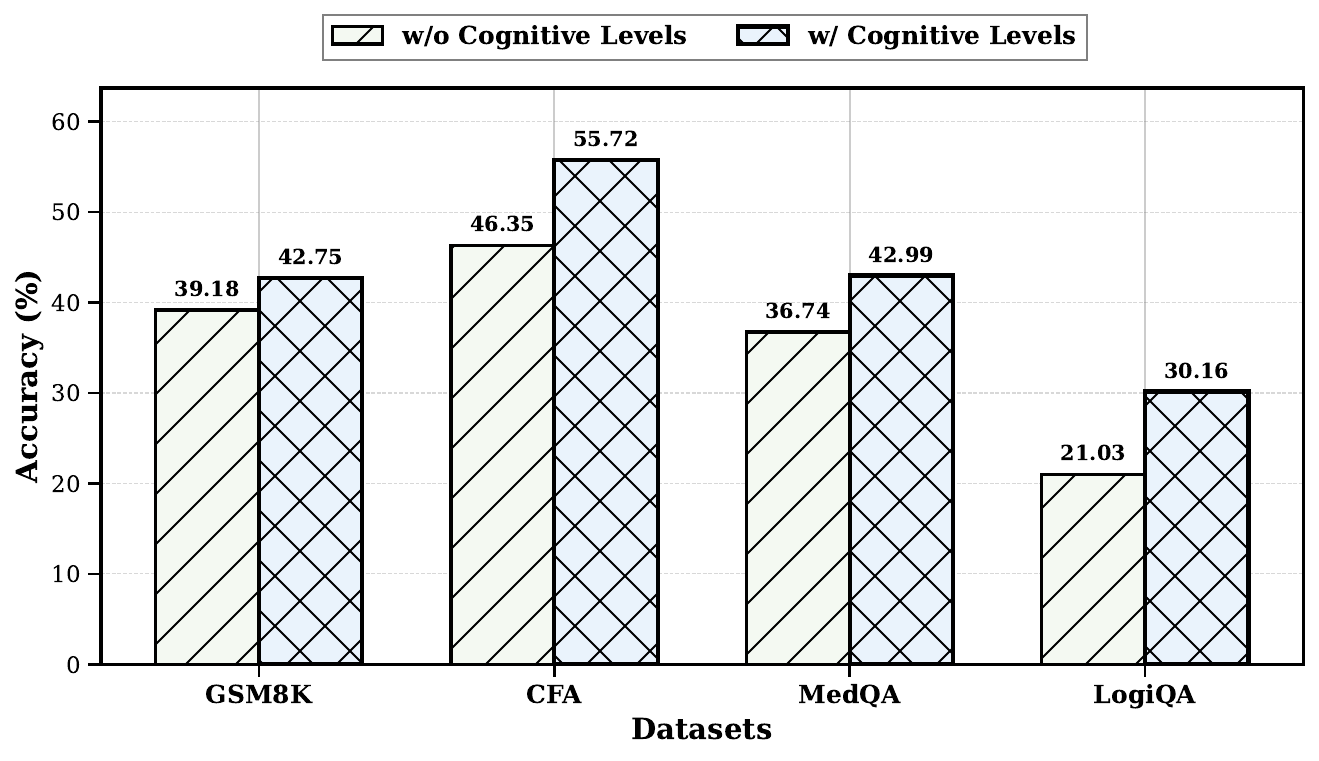}
    \caption{Ablation study evaluating the effectiveness of cognitive level integration in instruction generation.}
    \vspace{-10pt}
    \label{fig:ablation_cognitive}
\end{figure}

\noindent\textbf{Effectiveness of
Cognitive Level Instruction Generation.} To evaluate the impact of incorporating diverse cognitive levels in instruction generation, we compare two configurations using \texttt{Mistral-7B} across four datasets: (1) \textit{w/o Cognitive Levels}, which directly prompts the model to generate instructions by providing keywords without any cognitive complexity guidance, and (2) \textit{w/ Cognitive Levels}, our complete framework that systematically incorporates Bloom's Taxonomy to ensure instructions span different cognitive complexities. Figure~\ref{fig:ablation_cognitive} demonstrates the effectiveness of cognitive-level integration. The baseline approach without cognitive levels relies on simple keyword pairing, allowing the model to generate instructions without explicit guidance on the depth of reasoning required. In contrast, our cognitive-level framework explicitly structures instruction generation around Bloom's Taxonomy, ensuring a balanced distribution across different thinking processes, from basic recall to higher order analysis and creation.

\begin{figure}[h!]
    \centering
    \vspace{-10pt}
    \includegraphics[width=0.48\textwidth]{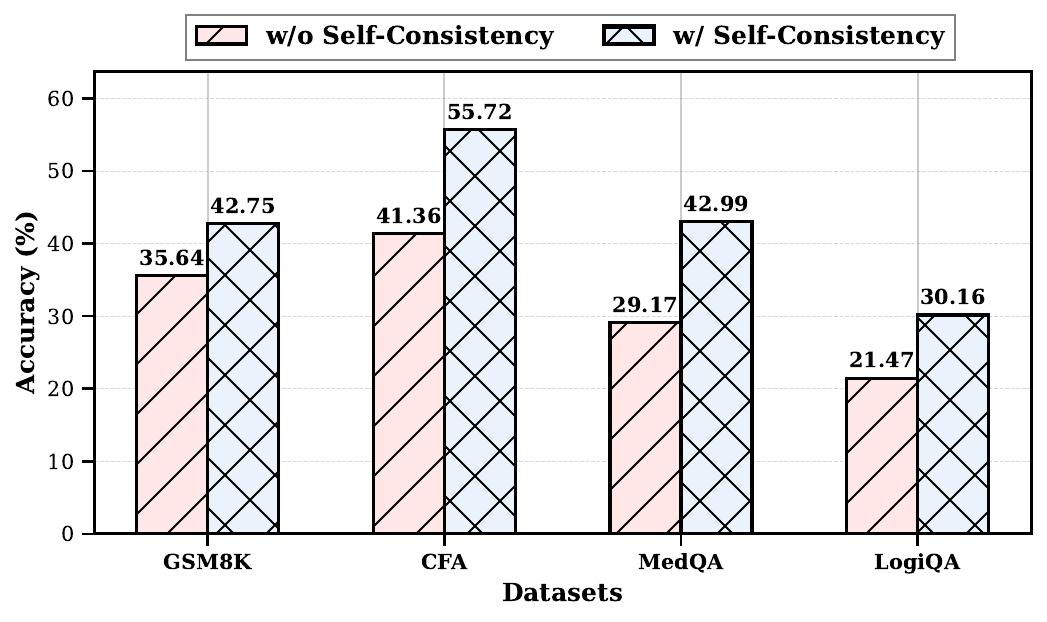}
    \caption{Ablation study comparing model performance with and without self-consistency filtering.}
    \vspace{-10pt}
    \label{fig:ablation_consistency}
\end{figure}

\noindent\textbf{Effectiveness of Self-Consistency Approach.} Our analysis compares performance with and without the self-consistency response filtering component described in Eq.~\ref{eq:vote_function}. The results, presented in Figure~\ref{fig:ablation_consistency}, demonstrate substantial and consistent improvements across all evaluated domains. The substantial improvements across all datasets confirm that response consistency serves as an effective proxy for instruction quality, enabling our framework to automatically identify and retain only the most reliable instruction-response pairs for downstream fine-tuning.

\label{subsec:main_results}

\section{Conclusion}
\label{sec:conclusion}
This paper introduces \textsc{DS\textsuperscript{2}-Instruct}, a zero-shot framework for generating high-quality, domain-specific instruction datasets without human supervision or seed examples. Our approach generates task-informed keywords for domain coverage, applies Bloom's Taxonomy for cognitive diversity, and uses self-consistency validation for quality filtering. Extensive experiments across seven specialized domains demonstrate that models fine-tuned on \frameworkname\xspace generated data substantially outperform existing synthetic instruction methods. Our ablation studies validate the necessity of each component, while scalability experiments across different data sizes and model scales confirm the robustness and generalizability of our approach. These results highlight \frameworkname\xspace's effectiveness in enhancing domain-specific reasoning capabilities without requiring expensive human annotation.

\section*{Limitations}
We identify three primary limitations in our work. First, our framework relies on static knowledge bases for keyword expansion, which may introduce coverage bias toward well-documented topics while potentially overlooking emerging or niche subfields. Incorporating dynamic knowledge retrieval or domain-specific corpora could help address this limitation. Second, the output quality depends on the generator LLM. Any biases or knowledge gaps in the generator can propagate to the synthesized data. Third, we evaluate our method only on English benchmarks, so its multilingual performance remains untested.


\bibliography{anthology,custom}
\appendix
\section{Dataset Details}
\label{sec:dataset}

We evaluate our DS$^2$-INSTRUCT framework across seven challenging domain-specific benchmarks spanning mathematics, finance, science, logical reasoning, and biomedicine. These benchmarks were selected to test diverse reasoning capabilities and domain-specific knowledge across multiple disciplines.

\begin{table*}[ht]
\centering
\caption{Overview of benchmark datasets used for evaluation.}
\label{tab:benchmark_overview}
\scalebox{0.8}{
\begin{tabular}{lcccc}
\toprule
\textbf{Dataset} & \textbf{Domain} & \textbf{Test Size} & \textbf{Question Type} & \textbf{Evaluation Metric} \\
\midrule
CFA~\cite{xie2023pixiu} & Finance & 1,506 & Multiple Choice & Accuracy \\
PubMedQA~\cite{jin2019pubmedqa} & Biomedical & 1,000 & Yes/No/Maybe & Accuracy \\
MedQA~\cite{jin2021disease} & Medical & 1,273 & Multiple Choice & Accuracy \\
GPQA~\cite{rein2024gpqa} & Graduate Science & 448 & Multiple Choice & Accuracy \\
LogiQA~\cite{liu2020logiqa} & Logical Reasoning & 651 & Multiple Choice & Accuracy \\
GSM8K~\cite{cobbe2021gsm8k} & Mathematics & 1,319 & Problem-Solving & Exact Match \\
MATH~\cite{hendrycks2021measuring} & Competition Math & 5,000 & Problem-Solving & Exact Match \\
\bottomrule
\end{tabular}
}
\end{table*}

The datasets encompass the following domains and characteristics:

\begin{itemize}
[leftmargin=*,itemsep=0.2pt,topsep=0.5pt]
    \item \textbf{CFA}~\cite{xie2023pixiu}: Chartered Financial Analyst exam questions testing asset valuation, portfolio management, wealth planning, and ethical standards. Questions require fundamental knowledge understanding, quantitative analysis, and practical application of financial concepts.
    
    \item \textbf{PubMedQA}~\cite{jin2019pubmedqa}: Biomedical research questions requiring yes/no/maybe answers based on corresponding abstracts from PubMed. This task demands strong comprehension and interpretation of biomedical literature and scientific reasoning.
    
    \item \textbf{MedQA}~\cite{jin2021disease}: Medical board exam questions testing professional-level knowledge across physiology, pathology, pharmacology, and clinical reasoning. Questions simulate real medical licensing examinations.
    
    \item \textbf{GPQA}~\cite{rein2024gpqa}: Graduate-level questions in Physics, Chemistry, and Biology requiring deep subject-matter knowledge, complex reasoning, calculation, and synthesis of advanced scientific concepts.
    
    \item \textbf{LogiQA}~\cite{liu2020logiqa}: Logical reasoning problems with multiple-choice questions that test the ability to understand logical structures, identify fallacies, and make valid inferences from given premises.
    
    \item \textbf{GSM8K}~\cite{cobbe2021gsm8k}: Grade school math word problems involving basic arithmetic, algebra, and geometry. Problems require step-by-step reasoning and mathematical problem-solving skills.
    
    \item \textbf{MATH}~\cite{hendrycks2021measuring}: Competition-level mathematics problems covering algebra, geometry, number theory, and calculus. These problems require sophisticated mathematical reasoning and step-by-step solution derivation.
\end{itemize}

\section{Additional Data Statistics}
Table~\ref{tab:self_instruct_statistics} shows statistics of data generated by Self-Instruct \cite{wang-etal-2023-self-instruct}.
\label{sec:statistics}
\begin{table*}[ht]
\centering
\caption{Statistical analysis of Self-Instruct generated datasets across different domains, including verb-noun (V-N) pair diversity metrics.}
\scalebox{0.9}{
  \begin{tabular}{lcccccccc}
  \toprule
  \textbf{Metric} & \textbf{CFA} & \textbf{PubMedQA} & \textbf{MedQA} & \textbf{GPQA} & \textbf{LogiQA} & \textbf{GSM8K} & \textbf{MATH} \\
  \midrule
  Avg. Length (words) $\uparrow$ & 85.65 & 53.33 & 91.92 & 41.98 & 79.91 & 46.39 & 37.69 \\
  Unique Verb-Noun Pairs $\uparrow$ & 505 & 45 & 333 & 287 & 196 & 117 & 68 \\
  Avg. Occurrences $\downarrow$ & 28.56 & 63.60 & 12.36 & 3.96 & 18.60 & 27.73 & 8.00 \\
  Std. Dev. Occurrences $\downarrow$ & 160.31 & 129.81 & 49.70 & 8.66 & 88.05 & 115.53 & 8.77 \\
  \bottomrule
  \end{tabular}
  \label{tab:self_instruct_statistics}
}
\end{table*}

\section{Implementation Details}
\label{sec:implementation}
For fair comparison across all methods, we generate 6,000 instruction-response pairs for each baseline and task. Our \textsc{DS\textsuperscript{2}-Instruct} framework begins with 50 initial seed keywords and performs bi-directional expansion, generating 5 new keywords in each direction per iteration over 100 total iterations. 

\subsection{Training Configuration}
We employ LoRA for efficient fine-tuning of pre-trained LLMs. Table~\ref{tab:training_config} presents the complete training hyperparameters used across all experiments.

\begin{table}[ht]
\centering
\caption{Fine-tuning hyperparameters for LoRA adaptation.}
\scalebox{0.95}{
\begin{tabular}{lc}
\toprule
\textbf{Hyperparameter} & \textbf{Value} \\
\midrule
LoRA Rank ($r$) & 8 \\
LoRA Alpha ($\alpha$) & 16 \\
Learning Rate & 2e-5 \\
Training Epochs & 5 \\
Batch Size & 16 \\
Weight Decay & 0.01 \\
Optimizer & AdamW \\
\bottomrule
\end{tabular}
}
\label{tab:training_config}
\end{table}

\subsection{Data Generation Configuration}
Our data generation pipeline integrates Retrieval-augmented keywords expansion and self-consistency filtering to ensure high-quality instruction-response pairs. 

\noindent\textbf{Keyword Generation Configuration.} Our keyword expansion strategy initializes with 50 seed keywords and generates 5 new keywords in each direction (forward and backward) per iteration. The expansion process runs for 100 iterations. For retrieval-augmented generation, we construct a diverse corpus by sampling from five complementary datasets in the Pile~\cite{gao2020pile}: ArXiv, FreeLaw, StackExchange, Wikipedia, and Github. This multidomain approach ensures comprehensive coverage across different writing styles and domain-specific terminology. We employ BM25~\cite{robertson2009probabilistic} as our ranking function. For each retrieval query, we retrieve the top-5 most relevant documents from the corpus. 

\noindent\textbf{Self-Consistency Filtering.} To ensure response quality, we apply self-consistency filtering with a threshold $\tau=3/5$. For each generated instruction, we sample $N=5$ response generations with temperature 0.7 and select responses that achieve majority agreement. The maximum generation length is set to 2048.

\section{Detailed Analysis of Cognitive Levels}
\label{sec:casestudy}
\subsection{Case Study}
To illustrate how \frameworkname\xspace generates cognitively diverse instructions across different levels of Bloom's Taxonomy, we present a detailed case study from the logical reasoning domain. We demonstrate how our framework systematically creates questions ranging from basic recall to complex creative tasks, all centered on domain-specific concepts. Table~\ref{tab:casestudy_examples} presents these generated instruction-response pairs, showcasing the cognitive diversity achieved by our approach.

\begin{table*}[htbp]
\centering
\small
\begin{tabular}{p{2.5cm}p{13cm}}
\toprule
\textbf{Cognitive Level} & \textbf{Generated Instruction} \\
\midrule
\textcolor{red!70!black}{\textbf{Remembering}} & 
What is the definition of satisfiability in the context of logic and problem-solving?

\textbf{Options:} A) The ability to prove that a statement is true in all possible scenarios. B) The process of finding a solution that meets all specified constraints. C) The method of determining the truth value of a logical expression. D) The technique of simplifying complex logical expressions to their basic components. \\
\midrule
\textcolor{orange!70!black}{\textbf{Understanding}} & 
Which statement best describes the roles of formal proof techniques and symbolic logic in constructing logical arguments?

\textbf{Options:} A) Formal proof techniques and symbolic logic are both essential tools in constructing logical arguments, each offering unique advantages in terms of clarity and rigor. B) Symbolic logic is the preferred method for constructing logical arguments, as it eliminates the need for formal proofs. C) Formal proof techniques are less precise than symbolic logic, but they are faster and easier to use in logical arguments. D) Neither formal proof techniques nor symbolic logic provide significant value in constructing logical arguments. \\
\midrule
\textcolor{green!70!black}{\textbf{Applying}} & 
In a logical reasoning test, you are given the statements: `If the system is secure, then the data is protected' (P $\rightarrow$ Q) and `The data is not protected' ($\neg$Q). Using modus tollens, what can you conclude about the security of the system?

\textbf{Options:} A) The system is secure. B) The data is protected. C) The system is not secure. D) The system's security status cannot be determined from the given statements. \\
\midrule
\textcolor{blue!70!black}{\textbf{Analyzing}} & 
In the context of logical reasoning, compare and contrast the application of ``basic set operations'' and ``verified observation'' in constructing a valid argument. How do these concepts interact when evaluating the strength and validity of an argument?

\textbf{Options:} A) Basic set operations are used to verify observations, ensuring that the logical structure of an argument is sound. B) Verified observation is a prerequisite for applying basic set operations, as it provides the necessary empirical data. C) Basic set operations and verified observation are independent processes that do not influence each other in logical reasoning. D) Basic set operations can be used to manipulate sets of verified observations to test the consistency and validity of an argument.\\
\midrule
\textcolor{purple!70!black}{\textbf{Evaluating}} & 
Evaluate the following statement: ``All bachelors are unmarried men, which is a truth known a priori.'' How does this statement relate to the concept of synthetic a priori knowledge, and what is the nature of the conclusion drawn?

\textbf{Options:} A) The statement is an example of analytic a priori knowledge, not synthetic a priori, because the predicate `unmarried' is contained within the subject `bachelor.' B) The statement is an example of synthetic a priori knowledge because it combines a priori reasoning with empirical facts about bachelors. C) The statement is neither analytic nor synthetic a priori because it relies on empirical observation to be true. D) The statement is an example of synthetic a posteriori knowledge because it requires empirical verification to confirm that all bachelors are indeed unmarried.\\
\midrule
\textcolor{brown!70!black}{\textbf{Creating}} & 
Design a logical reasoning problem where two events, A and B, always occur together (constant conjunction), but it is not clear whether one causes the other. Present a scenario and ask students to identify which additional information would best help determine if there is a causal relationship between A and B, or if their conjunction is coincidental.

\textbf{Example Scenario:} In a small town, every time the town clock strikes 12, a flock of birds flies over the town square. This has been observed consistently for the past year.

\textbf{Options:} A) The type of birds observed flying over the town square. B) The time of day when the birds are observed flying over the town square. C) Whether the birds still fly over the town square when the clock is temporarily stopped. D) The weather conditions when the birds are observed flying over the town square.\\
\bottomrule
\end{tabular}
\caption{Case study demonstrating \frameworkname\xspace's generation of cognitively diverse questions across all six levels of Bloom's Taxonomy for the logical reasoning domain.}
\label{tab:casestudy_examples}
\end{table*}

\subsection{Bloom Level Distribution Analysis}

To systematically evaluate cognitive demands across reasoning benchmarks, we randomly sampled 500 samples and conducted human annotation following Bloom's Taxonomy classification guidelines. Table~\ref{tab:bloom_comparison} presents distributions across GPQA, GSM8K, and MedQA, revealing distinct cognitive profiles.

\begin{table*}[t]
\centering
\begin{tabular}{lccc}
\toprule
\textbf{Bloom's Taxonomy Level} & \textbf{GPQA (\%)} & \textbf{GSM8K (\%)} & \textbf{MedQA (\%)} \\
\midrule
\textcolor{red!70!black}{\textbf{Remembering}} & 5.5 & 9.2 & 7.8 \\
\textcolor{orange!70!black}{\textbf{Understanding}} & 36.0 & 16.0 & 18.3 \\
\textcolor{green!70!black}{\textbf{Applying}} & 20.8 & 20.4 & 32.9 \\
\textcolor{blue!70!black}{\textbf{Analyzing}} & 24.0 & 25.2 & 23.3 \\
\textcolor{purple!70!black}{\textbf{Evaluating}} & 9.7 & 15.8 & 10.4 \\
\textcolor{brown!70!black}{\textbf{Creating}} & 4.0 & 13.4 & 7.3 \\
\midrule
\multicolumn{4}{c}{\textit{Aggregate Cognitive Categories}} \\
\midrule
\textbf{Lower-Order (Remember + Understand)} & 41.5 & 25.2 & 26.1 \\
\textbf{Middle-Order (Apply + Analyze)} & 44.8 & 45.6 & 56.2 \\
\textbf{Higher-Order (Evaluate + Create)} & 13.7 & 29.2 & 17.7 \\
\bottomrule
\end{tabular}
\caption{Bloom's Taxonomy distribution across benchmarks based on human annotation of 500 randomly sampled questions. GPQA emphasizes conceptual understanding (36.0\%), GSM8K shows balanced higher-order demands (29.2\%), and MedQA concentrates on procedural application (32.9\%).}
\label{tab:bloom_comparison}
\end{table*}

\section{Prompt Templates}
\label{sec:prompts}

\begin{tcolorbox}[
   colback=white,
   colframe=blue!50!black,
   colbacktitle=blue!50!black,
   coltitle=white,
   fonttitle=\small\bfseries,
   fontupper=\small,
   title=Cognitive Levels for Instruction Generation,
   breakable,
   arc=1mm,
   boxrule=1.2pt
]
\textcolor{red!70!black}{\textbf{Remembering}}: Create instructions that emphasize recall of factual knowledge, definitions, basic concepts, recognition tasks, and core terminology related to the keyword.

\textcolor{orange!70!black}{\textbf{Understanding}}: Design instructions that require conceptual understanding, explanation of relationships, interpretation, illustrative examples, and meaningful comparisons involving the keyword.

\textcolor{green!70!black}{\textbf{Applying}}: Formulate instructions that demand practical use of methods, implementation of procedures, execution of calculations, and real-world application of the keyword.

\textcolor{blue!70!black}{\textbf{Analyzing}}: Develop instructions that involve breaking down complex ideas, identifying patterns, examining relationships, and conducting comparative or structural analysis of the keyword.

\textcolor{purple!70!black}{\textbf{Evaluating}}: Construct instructions that involve critical judgment, validation of techniques, assessment of alternatives, justification of decisions, and critique of methods related to the keyword.

\textcolor{brown!70!black}{\textbf{Creating}}: Design instructions that foster original thinking, synthesis of ideas, problem innovation, creative design, and novel applications of the keyword.
\end{tcolorbox}

\subsection{Task-Specific Initial Keyword Generation Prompts}

\begin{tcolorbox}[
   colback=blue!5,
   colframe=blue!50!black,
   colbacktitle=blue!50!black,
   coltitle=white,
   fonttitle=\small\bfseries,
   fontupper=\small,
   title=CFA (Finance) - Initial Keywords,
   breakable,
   arc=1mm,
   boxrule=1.2pt
]
\textbf{Task Context}: You are an expert in Finance and CFA.

\textbf{Task Description}: Your task is to answer CFA exam questions in a multi-choice form, you should select the correct answer choice (e.g., A, B, C). These questions are about asset valuation, applying investment tools and concepts to analyze various investments, portfolio management, wealth planning, ethical and professional standards. It requires skills about Fundamental Knowledge Understanding, Quantitative Analysis and Calculations, Application and Analysis, etc.

\textbf{Instructions}: Generate 50 core keywords that represent the most essential concepts for this task.

\textbf{Requirements}:
\begin{itemize}
    \item List exactly 50 core concepts separated by commas
    \item Use underscores for multi-word concepts (e.g., asset\_valuation)
    \item Single words are acceptable (e.g., portfolio)
    \item Provide only the comma-separated list without any other text
\end{itemize}

\textbf{Core Keywords}:
\end{tcolorbox}

\begin{tcolorbox}[
   colback=blue!5,
   colframe=blue!50!black,
   colbacktitle=blue!50!black,
   coltitle=white,
   fonttitle=\small\bfseries,
   fontupper=\small,
   title=GSM8K (Mathematics) - Initial Keywords,
   breakable,
   arc=1mm,
   boxrule=1.2pt
]
\textbf{Task Context}: You are an expert in Mathematics.

\textbf{Task Description}: You are given a math word problem involving basic arithmetic, algebra, or geometry. Your task is to carefully read the problem and provide a step-by-step solution.

\textbf{Instructions}: Generate 50 core keywords that represent the most essential concepts for this task.

\textbf{Requirements}:
\begin{itemize}
    \item List exactly 50 core concepts separated by commas
    \item Use underscores for multi-word concepts (e.g., word\_problem)
    \item Single words are acceptable (e.g., addition)
    \item Provide only the comma-separated list without any other text
\end{itemize}

\textbf{Core Keywords}:
\end{tcolorbox}

\begin{tcolorbox}[
   colback=blue!5,
   colframe=blue!50!black,
   colbacktitle=blue!50!black,
   coltitle=white,
   fonttitle=\small\bfseries,
   fontupper=\small,
   title=PubMedQA (Biomedical) - Initial Keywords,
   breakable,
   arc=1mm,
   boxrule=1.2pt
]
\textbf{Task Context}: You are an expert in Biomedical Research.

\textbf{Task Description}: Your task is to answer biomedical research questions based on corresponding scientific abstracts. You must determine whether the answer to the question is "yes," "no," or "maybe" by analyzing the provided scientific text. This requires strong comprehension and interpretation of biomedical literature.

\textbf{Instructions}: Generate 50 core keywords that represent the most essential concepts for this task.

\textbf{Requirements}:
\begin{itemize}
    \item List exactly 50 core concepts separated by commas
    \item Use underscores for multi-word concepts (e.g., clinical\_trial)
    \item Single words are acceptable (e.g., treatment)
    \item Provide only the comma-separated list without any other text
\end{itemize}

\textbf{Core Keywords}:
\end{tcolorbox}

\begin{tcolorbox}[
   colback=blue!5,
   colframe=blue!50!black,
   colbacktitle=blue!50!black,
   coltitle=white,
   fonttitle=\small\bfseries,
   fontupper=\small,
   title=LogiQA (Logical Reasoning) - Initial Keywords,
   breakable,
   arc=1mm,
   boxrule=1.2pt
]
\textbf{Task Context}: You are an expert in Logical Reasoning.

\textbf{Task Description}: Your task is to solve logical reasoning problems multiple-choice questions. This involves analyzing a given logical puzzle or argument and choosing the correct option from a set of multiple-choice answers. These questions test your ability to understand logical structures, identify fallacies, and make valid inferences.

\textbf{Instructions}: Generate 50 core keywords that represent the most essential concepts for this task.

\textbf{Requirements}:
\begin{itemize}
    \item List exactly 50 core concepts separated by commas
    \item Use underscores for multi-word concepts (e.g., logical\_fallacy)
    \item Single words are acceptable (e.g., premise)
    \item Provide only the comma-separated list without any other text
\end{itemize}

\textbf{Core Keywords}:
\end{tcolorbox}

\begin{tcolorbox}[
   colback=blue!5,
   colframe=blue!50!black,
   colbacktitle=blue!50!black,
   coltitle=white,
   fonttitle=\small\bfseries,
   fontupper=\small,
   title=GPQA (Graduate Science) - Initial Keywords,
   breakable,
   arc=1mm,
   boxrule=1.2pt
]
\textbf{Task Context}: You are an expert in Graduate-level Science.

\textbf{Task Description}: Your task is to answer challenging, graduate-level multiple-choice questions spanning Physics, Chemistry, and Biology, requiring deep subject-matter knowledge, complex reasoning, calculation, and synthesis of information.

\textbf{Instructions}: Generate 50 core keywords that represent the most essential concepts for this task.

\textbf{Requirements}:
\begin{itemize}
    \item List exactly 50 core concepts separated by commas
    \item Use underscores for multi-word concepts (e.g., quantum\_mechanics)
    \item Single words are acceptable (e.g., thermodynamics)
    \item Provide only the comma-separated list without any other text
\end{itemize}

\textbf{Core Keywords}:
\end{tcolorbox}

\begin{tcolorbox}[
   colback=blue!5,
   colframe=blue!50!black,
   colbacktitle=blue!50!black,
   coltitle=white,
   fonttitle=\small\bfseries,
   fontupper=\small,
   title=MATH (Competition Mathematics) - Initial Keywords,
   breakable,
   arc=1mm,
   boxrule=1.2pt
]
\textbf{Task Context}: You are an expert in Competition Mathematics.

\textbf{Task Description}: You are given a challenging competition math problem. Your task is to carefully read the problem and provide a step-by-step solution for it.

\textbf{Instructions}: Generate 50 core keywords that represent the most essential concepts for this task.

\textbf{Requirements}:
\begin{itemize}
    \item List exactly 50 core concepts separated by commas
    \item Use underscores for multi-word concepts (e.g., number\_theory)
    \item Single words are acceptable (e.g., calculus)
    \item Provide only the comma-separated list without any other text
\end{itemize}

\textbf{Core Keywords}:
\end{tcolorbox}

\begin{tcolorbox}[
   colback=blue!5,
   colframe=blue!50!black,
   colbacktitle=blue!50!black,
   coltitle=white,
   fonttitle=\small\bfseries,
   fontupper=\small,
   title=MedQA (Medical Board Exams) - Initial Keywords,
   breakable,
   arc=1mm,
   boxrule=1.2pt
]
\textbf{Task Context}: You are an expert in Medical Knowledge.

\textbf{Task Description}: Your task is to answer medical questions collected from the professional medical board exams. These questions test professional-level knowledge across a broad range of medical domains, including physiology, pathology, pharmacology, and clinical reasoning. You should select the correct answer choice (e.g., A, B, C, or D).

\textbf{Instructions}: Generate 50 core keywords that represent the most essential concepts for this task.

\textbf{Requirements}:
\begin{itemize}
    \item List exactly 50 core concepts separated by commas
    \item Use underscores for multi-word concepts (e.g., differential\_diagnosis)
    \item Single words are acceptable (e.g., pharmacology)
    \item Provide only the comma-separated list without any other text
\end{itemize}

\textbf{Core Keywords}:
\end{tcolorbox}

\subsection{Task-Specific Instruction Generation Templates}

\begin{tcolorbox}[
   colback=blue!5,
   colframe=blue!50!black,
   colbacktitle=blue!50!black,
   coltitle=white,
   fonttitle=\small\bfseries,
   fontupper=\small,
   title=Instruction Generation Template,
   breakable,
   arc=1mm,
   boxrule=1.2pt
]
\textbf{Task Description}: [TASK\_DESCRIPTION]

\textbf{Keyword}: [KEYWORD]

\textbf{Question Type}: [COGNITIVE\_LEVEL] - [COGNITIVE\_LEVEL\_DESCRIPTION]

Generate a high-quality question that precisely targets the keyword and question type described above. The question should be clear, unambiguous, and suitable for instruction tuning.

Directly output the question. Do not include the answer or any other text.

\textbf{Generated Question}:
\end{tcolorbox}

\subsection{Task-Specific Response Suffixes}
\begin{tcolorbox}[
   colback=yellow!5,
   colframe=yellow!60!black,
   colbacktitle=yellow!60!black,
   coltitle=black,
   fonttitle=\small\bfseries,
   fontupper=\small,
   title=Response Format Suffixes by Task,
   breakable,
   arc=1mm,
   boxrule=1.2pt
]

\textbf{Multiple-Choice Tasks (CFA, LogiQA, MedQA, GPQA):} 
Return exactly two lines and nothing else: 
\texttt{Reason: <1--3 sentence explanation>} 
\texttt{Answer: <A|B|C|D>}

\vspace{0.3em}
\textbf{PubMedQA:} 
Return exactly two lines and nothing else: 
\texttt{Reason: <1--3 sentence explanation>} 
\texttt{Answer: <yes|no|maybe>}

\vspace{0.3em}
\textbf{GSM8K:} 
Provide a step-by-step reasoning process and then write the final numerical answer on a new line in the format: \texttt{final answer: <answer>}.

\vspace{0.3em}
\textbf{MATH:} 
Provide a step-by-step reasoning process and then write the final answer in the LaTeX boxed tag: \texttt{\$\textbackslash boxed\{<answer>\}\$}.

\end{tcolorbox}
\subsection{Universal Expansion and Generation Templates}

\begin{tcolorbox}[
   colback=gray!5,
   colframe=gray!50!black,
   colbacktitle=gray!50!black,
   coltitle=white,
   fonttitle=\small\bfseries,
   fontupper=\small,
   title=Bi-Directional Keyword Expansion Template,
   breakable,
   arc=1mm,
   boxrule=1.2pt
]
\textbf{Task Context}: You are an expert in the domain related to: [Task description $T_{desc}$].

\textbf{Sample Keywords}: [Randomly sampled keywords from current pool].

\textbf{Instructions}: Based on the sample keywords, generate new concepts in two directions:

1. \textbf{Prerequisite Concepts}: What fundamental concepts, basic terminology, or foundational principles should learners understand BEFORE studying the sample keywords?

2. \textbf{Advanced Concepts}: What specialized subfields, cutting-edge developments, or expert-level topics BUILD UPON the sample keywords?

\textbf{Requirements}:
\begin{itemize}
    \item Generate 5 concepts for each direction
    \item Use underscores for multi-word concepts
    \item Ensure concepts are different from existing keywords
    \item Provide comma-separated lists
\end{itemize}
\end{tcolorbox}

\begin{tcolorbox}[
   colback=white,
   colframe=promptboxblue,
   colbacktitle=promptboxblue,
   coltitle=white,
   fonttitle=\small\bfseries,
   fontupper=\small,
   title=Retrieval-Augmented Keyword Extraction Template,
   breakable,
   arc=1mm,
   boxrule=1.2pt
]
\textbf{Task Context}: You are an expert in the domain related to: {\color{red!70!black}[Task description $T_{desc}$]}.

\textbf{Current Keywords}: {\color{red!70!black}[List of expanded keywords]}.

\textbf{Retrieved Passages}: The following passages from authoritative sources (ArXiv, FreeLaw, StackExchange, Wikipedia, Github) contain comprehensive domain knowledge:
{\color{blue!70!black}[Top-k retrieved passages]}

\textbf{Instructions}: Extract additional domain-specific keywords directly from the retrieved passages that are missing from the current list.

\textbf{Extracted Keywords}:
\end{tcolorbox}
\subsection{Zero-shot Evaluation Prompts}

\begin{tcolorbox}[
   colback=blue!5,
   colframe=blue!60!black,
   colbacktitle=blue!60!black,
   coltitle=white,
   fonttitle=\small\bfseries,
   fontupper=\small,
   title=Zero-shot Evaluation Template,
   breakable,
   arc=1mm,
   boxrule=1.2pt
]

\textbf{General Structure}:

{\color{red!70!black}[Task Description $T_{desc}$]}

\vspace{0.2em}

{\color{blue!70!black}[Question from benchmark]}

\vspace{0.2em}

{\color{orange!70!black}[Task-specific response format suffix]}

\vspace{0.5em}

\textbf{Example for Multiple-Choice Tasks}:

You are tasked with solving finance/medical/science questions.

\vspace{0.2em}

Question: [Test question with options A, B, C, D]

\vspace{0.2em}

Return exactly two lines and nothing else: \\
\texttt{Reason: <1--3 sentence explanation>} \\
\texttt{Answer: <A|B|C|D>}

\vspace{0.5em}

\textbf{Example for Math Tasks (GSM8K/MATH)}:

You are tasked with solving mathematical problems.

\vspace{0.2em}

Question: [Math problem]

\vspace{0.2em}

Provide a step-by-step reasoning process and write the final answer in the format: \texttt{final answer: <answer>} (for GSM8K) or \texttt{\$\textbackslash boxed\{<answer>\}\$} (for MATH).

\end{tcolorbox}

\section{Ethics Statement}
We recognize the importance of evaluating and mitigating potential biases in automatically generated instruction data. To maintain transparency, we provide comprehensive details about our methodology, data sources, and evaluation procedures to enable reproducibility and critical assessment by the research community. Privacy is preserved throughout our framework by relying exclusively on publicly available resources and avoiding the collection or processing of sensitive personal information. Additionally, we disclose that ChatGPT was used exclusively for minor grammatical improvements and formatting corrections in the final manuscript preparation, without contributing to the core research methodology, analysis, or conclusions presented in this work.

\end{document}